\pdfoutput=1

\newcommand{\myspadesuit}{\textsuperscript{$\spadesuit$}}

\newcommand{\mydclubsuit}{\textsuperscript{$\clubsuit$}}

\documentclass[11pt]{article}

\usepackage[]{acl}

\usepackage{times}
\usepackage{latexsym}
\usepackage{amsmath}
\usepackage{graphicx} 
\usepackage{booktabs} 
\usepackage{siunitx}
\usepackage{multirow}
\usepackage{xcolor}
\usepackage{amssymb}
\usepackage[T1]{fontenc}

\usepackage[utf8]{inputenc}

\usepackage{microtype}

\title{Mitigating Hallucinations in Large Vision-Language Models with Instruction Contrastive Decoding}

\author{Xintong Wang\myspadesuit \and Jingheng Pan\myspadesuit \and Liang Ding\mydclubsuit\thanks{~~Corresponding Authors.} \and Chris Biemann\myspadesuit$^*$ \\
        \myspadesuit Department of Informatics, Universität Hamburg \\ 
        \mydclubsuit The University of Sydney \\
        \myspadesuit\{xintong.wang, jingheng.pan, chris.biemann\}@uni-hamburg.de \\
        \mydclubsuit liangding.liam@gmail.com}

\begin{document}
\maketitle
\begin{abstract}
Large Vision-Language Models (LVLMs) are increasingly adept at generating contextually detailed and coherent responses from visual inputs. However, their application in multimodal decision-making and open-ended generation is hindered by a notable rate of hallucinations, where generated text inaccurately represents the visual contents. To address this issue, this paper introduces the Instruction Contrastive Decoding (ICD) method, a novel approach designed to reduce hallucinations during LVLM inference. Our method is inspired by our observation that what we call disturbance instructions significantly exacerbate hallucinations in multimodal fusion modules. ICD contrasts distributions from standard and instruction disturbance, thereby increasing alignment uncertainty and effectively subtracting hallucinated concepts from the original distribution. Through comprehensive experiments on discriminative benchmarks (POPE and MME) and a generative benchmark (LLaVa-Bench), we demonstrate that ICD significantly mitigates both object-level and attribute-level hallucinations. Moreover, our method not only addresses hallucinations but also significantly enhances the general perception and recognition capabilities of LVLMs.
\end{abstract}

\section{Introduction}
Recent research in large vision-language models (LVLMs) \cite{liu2023visual, liu2023improved, li2023blip} has seen remarkable progress, benefiting from the integration of advanced large language models (LLMs) \cite{achiam2023gpt,touvron2023llama, touvron2023llama2} known for their robust language generation and zero-shot transfer capabilities~\cite{Zhong2023CanCU,peng-etal-2023-towards}. In order to leverage off-the-shell LLMs, it is crucial to facilitate cross-modal alignment.  LLaVa \cite{liu2023visual} employs a linear projection approach, while BLIP-2 \cite{li2023blip} and InstructBLIP \cite{liu2023improved} narrow the modality gap using a Q-Former. Although LVLMs have shown promising outcomes, the issue of hallucination remains. This phenomenon occurs when the generated textual content, despite being fluent and coherent, does not accurately reflect the factual visual content.

The object hallucination was initially explored within the realm of image captioning \cite{rohrbach2018object}.
As LVLMs harness the sophisticated understanding and generative prowess of LLMs, the scope of hallucination extends beyond mere object existence. It now encompasses more complex elements such as attributes and relationships within the generated content.  Consequently, distinguishing discriminative hallucination and the non-hallucinatory portion in the generation has become pivotal in assessing the performance of LVLMs in terms of their fidelity to factual visual information.

The intertwined nature of modalities presents significant challenges in identifying the root causes of hallucinations in LVLMs. Research efforts have begun to uncover the primary contributors to LVLM hallucinations, including statistical biases \cite{you2023ferret}  encountered during the training process and excessive dependence on language priors \cite{yan2023overcoming, zhibo2023overcoming}. Additionally, multimodal misalignment has been identified as a key factor in the occurrence of hallucinations \cite{jiang2023hallucination, liu2023mitigating,wang2023can}. To address dataset bias, annotation enrichment techniques \cite{gunjal2023detecting, you2023ferret, zhai2023halle} have been introduced. Furthermore, to counteract the influence of language priors, post-processing strategies \cite{yin2023woodpecker, zhou2023analyzing} have been developed, along with comprehensive initiatives aimed at improving multimodal alignment through optimizing alignment with humans \cite{sun2023aligning, jiang2023hallucination}. While these interventions have proven to be effective in reducing hallucinations, they demand substantial human involvement and incur significant computational costs for additional training or the integration of supplementary modules.

In this work, we reveal that appending instructions with role prefixes to form disturbance instructions can significantly exacerbate hallucinations.
We hypothesize that identifying and subsequently detaching hallucination concepts from the original distribution could effectively reduce such hallucinations. Motivated by this insight, we introduce the \textbf{Instruction Contrastive Decoding} ($\mathcal{\bf ICD}$) method. This approach is novel in that it is training-free and agnostic to the underlying LVLMs. ICD differentiates between two distributions: one from the original instruction and another from the disturbance instruction within the multimodal alignment module. Utilizing their difference, we aim at suppressing hallucinations. Through comprehensive experiments on discrimination hallucination benchmarks such as POPE \cite{li2023evaluating} and MME hallucination sets \cite{fu2023mme}, as well as the generation hallucination benchmark LLaVa-Bench \cite{liu2023visual}, our method incorporating state-of-the-art LVLMs like miniGPT4 and InstructBLIP, demonstrates significant efficacy in mitigating hallucinations at both object and attribute levels. 
Furthermore, our approach consistently enhances performance across 14 general perception and recognition tasks within the full MME benchmark.

Our main \textbf{contributions} are as follows:
\begin{itemize}
\item We perform an in-depth analysis of how disturbance in instructions exacerbates hallucinations. This phenomenon is elucidated through statistical bias and language priors, offering a nuanced understanding of underlying causes.
\item  Drawing on these insights above, we introduce the ICD method. This novel strategy, which emphasizes initial highlight followed by de-emphasize of hallucination, effectively mitigates hallucinations during inference, by adjusting the distributions away from hallucinations that we elicit. 
\item Through extensive experimentation and analysis, we validate the effectiveness of our proposed ICD method across both discrimination and generation hallucination benchmarks, showcasing its robustness and versatility in enhancing LVLMs performance.
\end{itemize}

\section{Related Work}

\paragraph{Large Vision-Language Models.} The field of vision-language pre-training (VLP) \cite{radford2021learning, li2022blip, bao2022vlmo, wang2023image} and fine-tuning \cite{wang2023using, wiehe2022language, alayrac2022flamingo} have seen rapid advancements, propelled by the evolution of large language models (LLMs). As a result, large vision-language models (LVLMs) have emerged, leveraging the strengths of frozen LLMs while emphasizing the facilitating of multimodal alignment modules. Notably, models such as LLaVa and Qwen-VL \cite{bai2023qwen} adopt simple linear projections to achieve alignment, contrasting with BLIP-2 and miniGPT4 \cite{zhu2023minigpt}, which introduce a Q-Former. In further work, InstructBLIP integrates task-aware instructions, enriching the understanding of task-aware visual semantics. Our research builds upon these advancements in LVLMs, focusing on the impact of instruction disturbances. We explore how such disturbances increase the uncertainty in multimodal alignment, significantly contributing to the exacerbation of hallucinations.

\paragraph{Hallucination in VLMs.} 
Generation hallucinations have been extensively studied in the field of language modeling~\cite{tonmoy2024comprehensive,wen2024perception}.
Hallucination in VLMs manifests as detailed, fluent, and coherent responses that inaccurately reflect the visual context, including erroneous objects, attributes, and relations~\cite{liu2024survey, jing2023faithscore}. 
Various strategies have been proposed to curb hallucinations. Annotation enrichment techniques like M-HalDetect \cite{gunjal2023detecting} and GRIT \cite{you2023ferret}, as well as approaches such as HACL \cite{jiang2023hallucination} and LLaVA-RLHL \cite{liu2023mitigating}, seek to improve alignment with human instructions through additional annotations. Similarly, Woodpecker \cite{yin2023woodpecker} introduces post-processing aimed at mitigating biases from language priors. While these methods have shown promise in reducing hallucinations, they often require extensive data annotation, fine-tuning, and supplementary modules, complicating their implementation. In contrast, our method directly addresses hallucinations during inference. Additionally, \citet{leng2023mitigating} introduced a visual contrastive decoding (VCD) approach that contrasts with the distributions of distorted visual inputs, a concept that bears resemblance to our method. However, our ICD method suppresses hallucinations through disturbance instructions affecting multimodal alignment.

\section{Method}
\subsection{Inference in LVLMs}
\label{sec:LVLM}

\begin{figure*}[htbp] 
  \centering
  \includegraphics[width=1.0\textwidth]{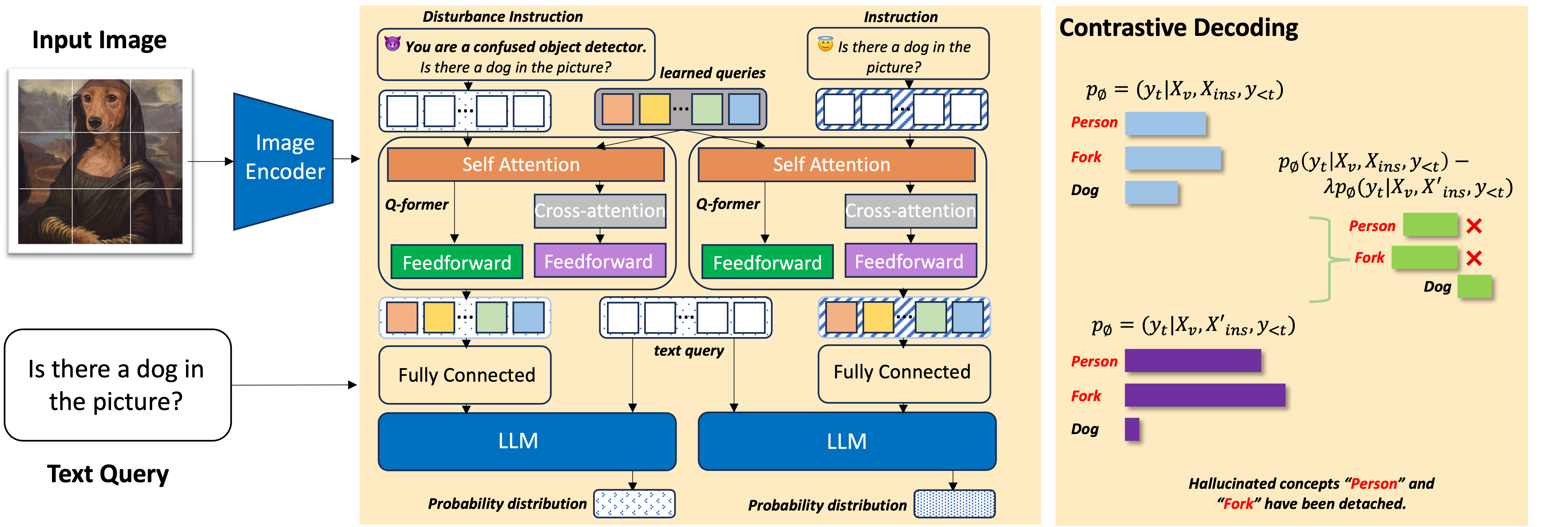} 
  \caption{\textbf{An illustration on inference framework and contrastive decoding process of ICD method.} At the core \textcolor{orange}{(middle orange box)}, the framework integrates a frozen image encoder, LLM, and query vectors \textcolor{gray}{(gray box)} within the Q-Former, focusing solely on adjusting the standard and disturbance instructions. The latter, exemplified by adding role prefixes like \textbf{\textcolor{purple}{`You are a confused object detector,'}} aims to increase multimodal alignment uncertainty. This results in two distinct distributions: one from the standard instruction and another influenced by the disturbance. The contrastive decoding method \textcolor{orange}{(right orange box)} highlights how disturbance instructions amplify hallucinated concepts (\textcolor{red}{`person and fork'}), which are then corrected by subtracting probabilities derived from the standard instruction, ensuring accurate recognition of the correct concept \textcolor{teal}{`dog'}.}
  \label{fig:pnprefix} 
\end{figure*}

Large Vision-Language Models (LVLMs) are comprised of three pivotal components: a visual encoder, a fusion module, and a language model. For processing an input image, a pre-trained visual encoder, such as ViT-L/14 from CLIP \cite{radford2021learning}, is employed to extract visual features, denoted as $\mathbf{X_{V}}$. The fusion module facilitates multimodal alignment. For instance, InstructBLIP introduces an instruction-aware querying transformer. Q-Former, a lightweight transformer architecture, utilizes $K$ learnable query vectors $\mathbf{Q_{K}}$ to refine the extraction of visual features, thereby enhancing multimodal alignment. It allows the instruction $\mathbf{X_{ins}}$ to interact with the query vectors, fostering the extraction of task-relevant image features:

{\small
\begin{equation}
Z_{V} = Q_{\theta}(X_{V}, Q_{K}, X_{ins}),
\end{equation}}where, $Z_V = Q_{\theta}(\cdot)$ represents the fused visual features, conditioned on the instructions. Given its sophistication and effectiveness in multimodal alignment, we advocate for the adoption of the instruction-aware Q-Former architecture.

For text queries $\mathbf{X_{q}}$, a large language model, parameterized by $\phi$, such as Vicuna \cite{chiang2023vicuna}, processes the query, leveraging the derived visual features to formulate responses:

{\small
\begin{equation}
Y_{R} = LLM_{\phi} (H_{V}, X_{ins}),
\end{equation}}where $H_{V} = g(Z_{V})$ is the transformation ensuring the same dimensionality as the word embedding of the language model. By default, the instruction is the same as text query for both Q-Former and LLM as $\mathbf{X_{ins}=X_{q}}$.

Mathematically, in the decoding phase, the response $\mathbf{R}$ can be defined as a sequence of length $\mathbf{L}$, sampled from a probability distribution:

{\small
\begin{equation}
p(Y_{R} | X_{V}, X_{q}) = \prod_{t=1}^{L} p_{\phi}(y_{t} | H_{V}, X_{q}, y_{<t}),
\label{eq:inference}
\end{equation}}where $y_{<t}$ represents the sequence of generated tokens up to the time steps $(t-1)$. In the decoding phase of LVLMs, hallucinations often emerge when probable tokens lack grounding in the visual context. \citet{jiang2023hallucination, liu2023mitigating} indicate that multimodal misalignment is a critical factor contributing to the generation of hallucinations. Thus, we conduct an in-depth analysis of the fusion module, specifically focusing on multimodal alignment. Our work first demonstrates that instructions within the multimodal alignment module can exacerbate hallucinations. To address this, we introduce instruction disturbance and propose an instruction contrastive decoding method, employing a \textbf{highlight and then detach} strategy.

\subsection{Instruction Can Amplify Hallucination}
Prior studies have attributed the occurrence of hallucinations in LVLMs to statistical biases within multimodal training datasets \cite{you2023ferret} and an over-reliance on language priors \cite{yan2023overcoming, zhibo2023overcoming}. Extending this line of observation, we introduce the concept of instruction disturbance in this section. A prefix appended to instructions affects multimodal alignment, thereby exacerbating statistical biases and the over-reliance on language priors.

\textbf{Introduction of instruction disturbance}: 
We introduce the concept of instruction disturbance, which entails appending a \textit{role prefix} to the original instructions delineated in Section~\ref{sec:LVLM}. This disturbance aims to modulate the multimodal alignment uncertainty within LVLMs. As illustrated in Figure~\ref{fig:pnprefix}, the base instruction \textit{“Describe this photo in detail”} is combined with learned query vectors in the Q-Former. To implement instruction disturbance, we append either \textit{positive} or \textit{negative} prefixes to the base instruction. Positive prefixes aim to increase the LVLM's confidence in multimodal alignment. Conversely, negative prefixes are designed to reduce the model's alignment confidence.

{\small
\begin{equation}
X_{ins} =
\begin{cases}
[X_d, X_q] & \text{\textit{if disturbance}} \\
X_q & \text{\textit{otherwise}}
\end{cases},
\label{eq:dis}
\end{equation}}where $X_{d}$ denotes the role prefix, and $X_{q}$ represents the original instruction. Through this method, we strategically influence the LVLM’s confidence level in multimodal alignment by either encouraging a more definitive understanding or introducing ambiguity.

\textbf{Instruction disturbance amplifies statistical biases and language priors:} Figure~\ref{fig:pnprefix} presents the response from InstructBLIP, revealing that the LVLMs generate hallucinated tokens such as “\textit{fork and person}.” To further explore this phenomenon, we undertake two specific analyses: the frequent hallucinated object occurrence and the co-occurrence of object hallucinations. Our study utilizes MSCOCO validation set \cite{lin2014microsoft}, a common dataset for LVLM pre-training, to perform hallucination detection across three distinct scenarios: the baseline LVLM, LVLM with a positive disturbance, and LVLM with a negative disturbance. Our analysis focuses on calculating the hallucination ratio, specifically identifying instances where the hallucinated objects are absent from the provided images.

\begin{figure}[htbp]
  \centering
  \includegraphics[width=\columnwidth]{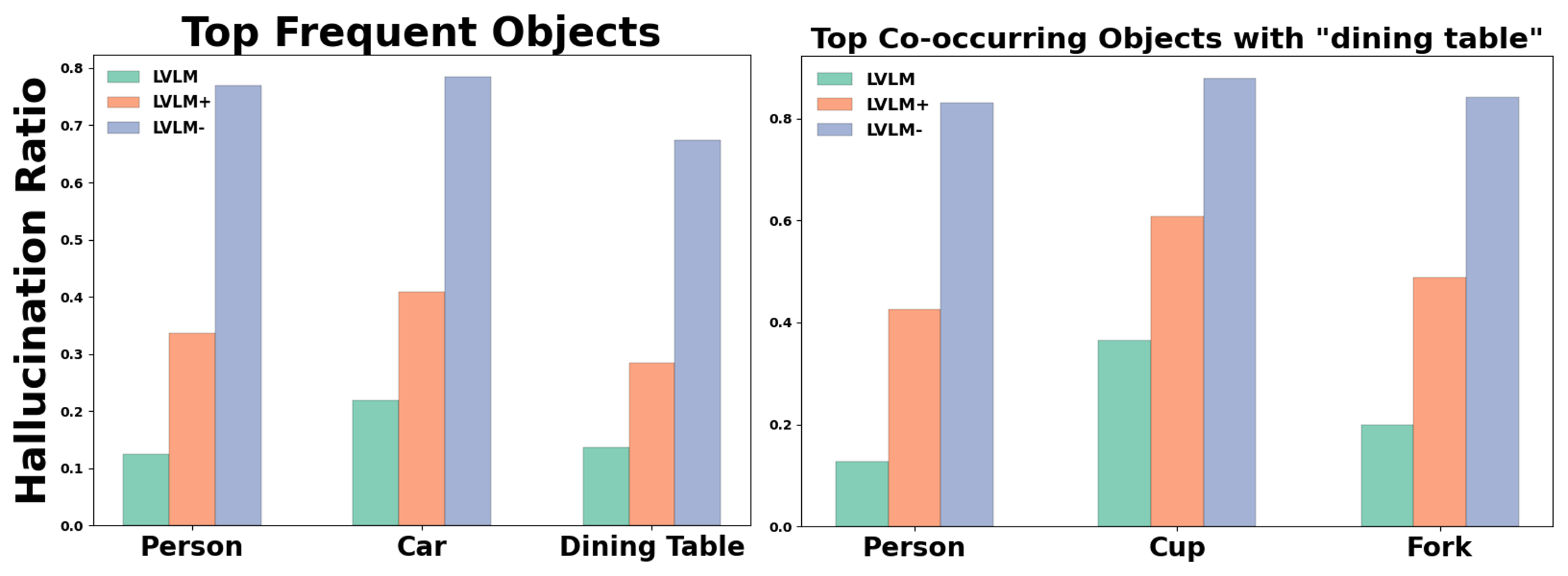}
  \caption{The left figure shows the \textbf{top frequent objects hallucination ratio} and the right depicts the \textbf{ratio of co-occurring object hallucinations} with \textit{dining table}.}
  \label{fig:bias} 
\end{figure}

Figure~\ref{fig:bias} demonstrates that introducing instruction disturbance significantly amplifies the occurrence of hallucinations. Under the influence of negative disturbance, LVLMs are more likely to hallucinate objects that frequently co-occur, such as “\textit{person and dining table},” and show an increased tendency to hallucinate objects that typically co-occur with those actually present in the image, for example, “\textit{fork and person}.” This suggests that instruction disturbances, whether positive or negative, intensify the hallucination effect, exacerbating the issues of imbalanced object distribution and correlation patterns inherent in the training dataset.


\subsection{Instruction Contrastive Decoding}
\subsubsection{Contrastive Decoding with Disturbance}
Our analysis reveals that instruction disturbances exacerbate hallucinations by increasing multimodal alignment uncertainty. This uncertainty predisposes LVLMs to more readily adopt biased co-occurrence concepts from pretraining datasets, as reflected in the learned query vectors. As these hallucinations accumulate, LVLMs increasingly over-rely on language priors. Notably, disturbances involving negative prefixes significantly intensify these hallucinations. We hypothesize that by initially emphasizing the probabilities of hallucinated concepts and subsequently detaching these from the original probability distribution, hallucinations may be reduced. Inspired by this insight, we introduce an Instruction Contrastive Decoding method (ICD) aimed at mitigating hallucinations during LVLM inference. 

Motivated by the language contrastive decoding \cite{sennrich2023mitigating} in reducing hallucinations within machine translation frameworks—where it prevents potentially accurate translations that, however, deviate from the desired target language—we adopt a similar approach to our model. Given the extraction of visual features $\mathbf{X_{V}}$ from the visual encoder and a textual query $\mathbf{X_{q}}$, our model calculates two distinct token distributions: one conditioned on the original instructions, and the other on instructions with disturbance $\mathbf{X_{d}}$ as Equation~\ref{eq:dis}. Contrary to the conventional approach of selecting the token that maximizes the probability, our strategy involves choosing the token that concurrently maximizes $p_{\phi}(y_{t} | X_{V}, X_{ins})$ and minimizes $p_{\phi}(y_{t} | X_{V}, X_{ins}')$, the latter representing the probability of tokens that are more likely to be hallucinations. To adjust the balance between these probabilities, we introduce a hyperparameter $\lambda$, which regulates the intensity of the contrastive penalty. Formally, this process is described as follows:

{\small
\begin{align}
p_{icd}(Y_{R} | X_{V}, X_{q}) &= \prod_{t=1}^{L} \Big( p_{\phi}(y_{t} | X_{V}, X_{ins}, y_{<t}) \nonumber \\
&\quad - \lambda p_{\phi}(y_{t} | X_{V}, X_{ins}', y_{<t}) \Big),
\label{eq:icd}
\end{align}
}where larger $\lambda$ indicates a more decisive penalty on the decision made by LVLMs with disturbances.

\subsubsection{Adaptive Plausibility Constrains}
The ICD objective is designed to favor tokens preferred by the LVLM output while imposing penalties on tokens influenced by instruction disturbances. However, this approach might inadvertently penalize accurate predictions—those tokens that, under both standard and disturbance instruction conditions, are confidently identified and are well-grounded in the visual context (such as objects, verbs, attributes, and relations) due to their simplicity and high likelihood. Conversely, it might erroneously reward tokens representing implausible concepts. To address this issue, we draw inspiration from adaptive plausibility constraints utilized in open-ended text generation \cite{li2022contrastive}. Consequently, we refine the ICD objective to incorporate an adaptive plausibility constraint:

{\small
\begin{align}
y_{t} &\sim softmax \Big( logit_{\phi}(y_{t} | X_{V}, X_{ins}, y_{<t}) \nonumber \\
&\quad - \lambda logit_{\phi}(y_{t} | X_{V}, X_{ins}', y_{<t}) \Big) \nonumber \\
&\quad \textit{subject to } y_{t} \in \mathcal{V}_{head}(y_{<t})
\label{eq:icdc}
\end{align}
\begin{align}
\mathcal{V}_{head}(y_{<t}) &= \biggl\{ y_t \in \mathcal{V} : 
p_{\phi}(y_t | X_{V}, X_{ins}, y_{<t}) \nonumber \\
&\qquad \geq \alpha \max_{token} p_{\phi}(token | X_{V}, X_{ins}, y_{<t}) \biggr\},
\label{eq:myequation}
\end{align}
}here, $\alpha$ acts as a pivotal hyperparameter that modulates the truncation of the probability distribution, effectively tailoring the LVLM's response to its confidence level. This is particularly crucial for mitigating the influence of implausible tokens, especially when LVLMs exhibit high confidence and are accurately anchored in visual semantics.



ICD serves as a self-corrective mechanism, which successfully identifies hallucinations in LVLMs and then de-emphasizes them through contrastive decoding. Moreover, the integration of adaptive plausibility constraints further hones the contrastive distribution by considering the confidence levels of LVLMs, thereby narrowing the decision-making process to a more reliable candidate pool. This method not only significantly reduces hallucinations within LVLMs but also curtails the generation of implausible tokens, showcasing the efficacy of our proposed method in enhancing model reliability and output validity.

\section{Experiment}
\label{experiment}
In this section, we explore the evaluation of our ICD method for mitigating hallucinations. Our examination is twofold: firstly, through the lens of hallucination discrimination, and secondly, via the generation of non-hallucinatory content. More precisely, we assess the efficacy of ICD in alleviating object-level hallucination symptoms utilizing the POPE benchmark. Furthermore, we extend our analysis to include both object and attribute-level symptoms through the MME benchmark. Finally, the performance of our method in generating non-hallucinatory content is evaluated using the LLaVa-Bench dataset.

\subsection{Experimental Settings}
\begin{table*}[t]
\centering
\small 
\setlength\tabcolsep{3pt} 
\begin{tabular}{
    l
    l
    l
    S[table-format=2.2]
    S[table-format=2.2]
    S[table-format=2.2]
    S[table-format=2.2]
    S[table-format=2.2]
    S[table-format=2.2]
    S[table-format=2.2]
    S[table-format=2.2]
    S[table-format=2.2]
}
\toprule
\textbf{Dataset} & \textbf{Setting} & \textbf{Method} & \multicolumn{4}{c}{\textbf{miniGPT4 Backbone}} & \multicolumn{4}{c}{\textbf{InstructBLIP Backbone}} \\
\cmidrule(lr){4-7} \cmidrule(lr){8-11}
 &  &  & {\textbf{Accuracy}} & {\textbf{Precision}} & {\textbf{Recall}} & {\textbf{F1 Score}} & {\textbf{Accuracy}} & {\textbf{Precision}} & {\textbf{Recall}} & {\textbf{F1 Score}} \\
\midrule
\multirow{9}{*}{MSCOCO} & \multirow{3}{*}{Random} & \textit{default} & 67.04 & 69.06 & 66.54 & 67.77 & 80.71 & 81.67 & 79.19 & 80.41 \\
 &  & \textit{+vcd} & 69.60 & 72.76 & 66.73 & 69.62 & 84.53 & 88.55 & 79.32 & 83.68 \\
 &  & \textit{+icd} & \textbf{73.51} & \textbf{74.36} & \textbf{76.87} & \textbf{75.60} & \textbf{86.43} & \textbf{92.01} & \textbf{80.73} & \textbf{85.61} \\
 \addlinespace 
  & \multirow{3}{*}{Popular} & \textit{default} & 60.89	& 61.34&	65.74&	63.46&	78.22&	77.87	&78.85&	78.36\\
 &  & \textit{+vcd} & 62,91 &	63,69&	64,81	&64,24	&81.47&	82.89&	79.32&	81.07 \\
 &  & \textit{+icd} & \textbf{67.61} &	\textbf{66.69} &	\textbf{76.87}	& \textbf{71.42} &	\textbf{82.93} &	\textbf{84.45} &	\textbf{80.73}	& \textbf{82.55} \\
 \addlinespace 
  & \multirow{3}{*}{Adversarial} & \textit{default} & 59.42&	59.64	&64.45&	61.95	&75.84&	74.30	&79.03&	76.59 \\
 &  & \textit{+vcd} & 62.07	& 62.15	& 66.76	&64.37	&79.56	&79.67	&79.39	&79.52 \\
 &  & \textit{+icd} & \textbf{64.36}	&\textbf{63.68}	&\textbf{75.11}	&\textbf{68.93}	&\textbf{80.87}	&\textbf{80.95}	&\textbf{80.73}	&\textbf{80.84} \\
 \midrule
 \multirow{9}{*}{A-OKVQA} & \multirow{3}{*}{Random} & \textit{default} & 64.79&	65.26&	65.73&	65.50&	80.91&	77.97	&86.16&	81.86 \\
 &  & \textit{+vcd} & 66.68	&66.47	&68.21	&67.33	&84.11	&82.21&	87.05&	84.56 \\
 &  & \textit{+icd} & \textbf{69.04}	&\textbf{68.50}	&\textbf{77.04}	&\textbf{72.52}	&\textbf{85.82}	&\textbf{83.80}	&\textbf{88.94}	&\textbf{86.29} \\
 \addlinespace 
  & \multirow{3}{*}{Popular} & \textit{default} & 60.75	&60.67&	68.84&	64.50&	76.19&	72.16&	85.28&	78.17 \\
 &  & \textit{+vcd} & \underline{62.22} &	\textbf{62.23}&	68,55&	65,24	&79.78	&76.00	&87.05	&81.15 \\
 &  & \textit{+icd} & \textbf{62.81}	& \underline{61.62}	&\textbf{75.78}	&\textbf{67.97}	&\textbf{81.64}	&\textbf{78.50}	&\textbf{88.77}	&\textbf{83.32} \\
 \addlinespace 
  & \multirow{3}{*}{Adversarial} & \textit{default} & 58,88&	58,56&	68,50&	63,14&	70.71&	65.91&	85.83&	75.56 \\
 &  & \textit{+vcd} & \underline{60.67} &	\textbf{60.56}&	68.47&	64.28&	\underline{74.33} &	\underline{69.46} &	86.87&	77.19 \\
 &  & \textit{+icd} & \textbf{60.71}&	\underline{59.27}	&\textbf{77.68}&	\textbf{67.24}	&\textbf{74.42}	&\textbf{70.24}	&\textbf{88.93}	&\textbf{78.48} \\
 \midrule
 \multirow{9}{*}{GQA} & \multirow{3}{*}{Random} & \textit{default} & 65.13	&65.38	&66.77	&66.07&	79.75&	77.14	&84.29&	80.56 \\
 &  & \textit{+vcd} & 67.08	&68.30	&69.04	&68.67&	83.69&	81.84&	\textbf{86.61}	&84.16\\
 &  & \textit{+icd} & \textbf{72.24}	&\textbf{75.08}	&\textbf{79.54}	&\textbf{77.24}	&\textbf{85.10}	&\textbf{84.21}&	\underline{86.40}	&\textbf{85.29} \\
 \addlinespace 
  & \multirow{3}{*}{Popular} & \textit{default} & 57.19& 	58.55& 	60.81& 	59.66& 	73.87& 	60.63& 	84.69& 	76.42 \\
 &  & \textit{+vcd} & \underline{62.14}	& \textbf{61.14}& 	72.26& 	66.24& 	\underline{78.57} & 	\underline{74.62}& 	\underline{86.61}& 	\underline{80.17} \\
 &  & \textit{+icd} & \textbf{62.84} & 	\underline{61.09}& 	\textbf{80.54} & 	\textbf{69.48} & 	\textbf{78.80}	& \textbf{75.15}	& \textbf{87.53}	& \textbf{80.87} \\
 \addlinespace 
  & \multirow{3}{*}{Adversarial} & \textit{default} & 56.75& 	56.26& 	67.99	& 61.57	& 70.56	& 66.12	& 84.33	& 74.12 \\
 &  & \textit{+vcd} & 57.78	& \underline{57.70}	& 69.82	& 63.18& 	\underline{75.08} & 	\underline{70.59}	& \underline{85.99} & \underline{77.53} \\
 &  & \textit{+icd} & \textbf{59.64}	& \textbf{58.21}	& \textbf{76.81}& 	\textbf{66.23}& 	\textbf{75.17}	& \textbf{70.59}& 	\textbf{86.27}& 	\textbf{77.65} \\
\bottomrule
\end{tabular}
\caption{\textbf{Results on discrimination hallucination benchmark POPE.} The default under methods denotes the standard decoding, whereas VCD represents visual contrastive decoding \cite{leng2023mitigating}, and ICD is our instruction contrastive decoding. The best performances within each setting are \textbf{bolded}. Comparable ($\pm 1.0$) but not the best performances between VCD and ICD methods are \underline{underlined}.}
\label{tab:POPE}
\end{table*}
\subsubsection{Datasets and Evaluation Metrics}
\textbf{POPE:} The Polling-based Object Probing Evaluation (POPE) stands as a popular benchmark in discerning hallucination at the object level. POPE employs a binary question-answering format, inquiring LVLMs to determine the presence or absence of a specified object within a given image. This benchmark is structured around three distinct subsets—MSCOCO, A-OKVQA \cite{schwenk2022okvqa}, and GQA \cite{hudson2019gqa}—each comprising 500 images alongside six questions per image. POPE introduces three settings within each subset: \textit{random} (selecting absent objects at random), \textit{popular} (choosing the most frequently occurring objects in the dataset as absent), and \textit{adversarial} (selecting absent objects that often co-occur with ground-truth objects). We adopt Accuracy, Precision, Recall, and F1 score as the evaluation metrics.

\textbf{MME:} MME benchmark serves as a comprehensive tool for assessing the capabilities of LVLMs across both perception and cognition, spanning a total of 14 tasks. Among these, tasks focusing on \textit{existence, count, position, and color} are specifically designed as hallucination discrimination benchmarks. These tasks aim to scrutinize both \textit{object-level} and \textit{attribute-level} hallucination symptoms. MME similarly utilizes a question-answering format to facilitate this evaluation. Consequently, task scores are reported as the evaluation metric for measuring performance.

\textbf{LLaVa-Bench:} The LLaVa-Bench is designed to quantify the extent of hallucinated content produced during the open-ended generation tasks performed by LVLMs. This benchmark encompasses a varied collection of 24 images, accompanied by 60 questions that cover a wide range of scenarios, including indoor and outdoor scenes, memes, paintings, and sketches. Unlike discriminative benchmarks, where accuracy serves as the evaluation metric, generative benchmarks, such as this, currently do not have well-established metrics specifically devised for the detailed analysis of hallucinations \cite{liu2024survey}. Therefore, we utilize case studies on this dataset as a means to qualitatively evaluate the effectiveness of our ICD method (see in appendix~\ref{sec:case}).


\begin{figure*}[htbp] 
  \centering
  \includegraphics[width=1\textwidth]{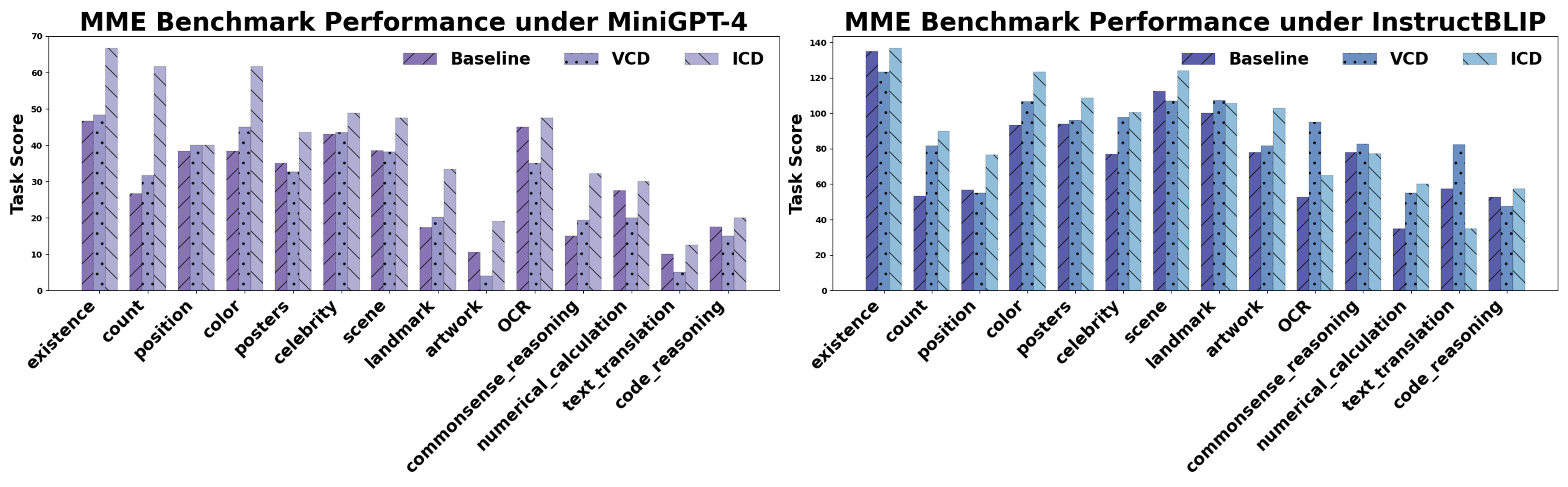} 
  \caption{\textbf{Performance on MME full benchmark}. The left figure in purple is the results based on miniGPT4, while the right figure in blue is the results based on InstructBLIP.}
  \label{fig:mme} 
\end{figure*}

\subsubsection{LVLM Baselines} 
We employ two state-of-the-art LVLMs as backbone frameworks. 
Specifically, we implement our ICD on InstructBLIP and miniGPT4, which utilize the Vicuna 7B as their underlying LLM and the sophisticated Q-Former architecture for fusion modules, respectively. Additionally, we explore the use of LLaVa-1.5 \cite{liu2023improved}, which incorporates linear projection for its fusion module alongside InstructBLIP, to identify optimal practices in applying the ICD method (see in appendix~\ref{sec:cd}). Finally, we compare our method against the visual contrastive decoding approach \cite{leng2023mitigating}, designed to mitigate hallucinations arising from visual uncertainties. We posit that our method, being LVLM-agnostic, can be conveniently integrated into various off-the-shelf LVLMs.

\subsection{Experimental Results}
\subsubsection{Results on POPE}
The experimental results on POPE, summarized in Table~\ref{tab:POPE}, demonstrate the efficacy of our instruction contrastive decoding method across three distinct subsets within the POPE benchmark—MSCOCO, A-OKVQA, and GQA settings. Notably, our ICD method consistently outperforms the foundational LVLMs, miniGPT4, and InstructBLIP. Specifically, the ICD method exceeds the performance of miniGPT4 and InstructBLIP, showing a substantial improvement of $\textbf{10.5\%}$ and $\textbf{6.0\%}$, respectively, across all metrics ($\textbf{7.0\%}$ in accuracy,  $\textbf{8.5\%}$ in precision, $\textbf{8.7\%}$ in recall, and $\textbf{7.9\%}$ in F1 score for both models). This significant enhancement as per four metrics on POPE underscores the effectiveness of our \textit{highlight and then detach} strategy. 

Furthermore, the progressive movement from \textit{random} to \textit{popular} and then to \textit{adversarial} settings reveals a marked decline in performance, highlighting the growing impact of statistical biases and language prior to contributing to hallucinations in LVLMs. Despite these challenges, our ICD method consistently demonstrates improvements across all settings, affirming our hypothesis that disturbance instruction exacerbates hallucinations by influencing multimodal alignment, thereby deepening errors rooted in statistical bias and over-reliance on language priors, which can be subtracted by contrastive decoding. Our method effectively mitigates these issues and object-level hallucinations.

In comparison to the VCD approach, our ICD method achieves an overall improvement of $\textbf{3.9\%}$. While the VCD method aims to ensure that the output distributions are closely aligned with visual inputs and compares distributions derived from distorted images, it requires additional processing to distort images via diffusion models \cite{ho2022classifier} and is sensitive to the choice of hyperparameters in its experimental setup \cite{leng2023mitigating}. Conversely, our ICD method offers a more straightforward and efficient solution, yielding superior results in an end-to-end manner.

\subsubsection{Results on MME}

\textbf{Results on MME Hallucination Subset:} The analysis of the POPE benchmark underscores the efficacy of our ICD method in mitigating object-level hallucination symptoms. Given that hallucinations can also manifest at the attribute level \cite{liu2024survey}, it becomes imperative to extend our investigation to these dimensions. To this end, we leverage the MME hallucination subset, which encompasses both object-level (\textit{existence and count tasks}) and attribute-level (\textit{position and color tasks}) benchmarks, to conduct a comprehensive evaluation of the ICD method.

\begin{table}[t]
\centering
\label{tab:MMEhal}
\resizebox{\columnwidth}{!}{%
\begin{tabular}{
    l
    l
    c
    c
    c
    c
    c
}
\toprule
\textbf{LVLM} & \textbf{Method} & \multicolumn{2}{c}{\textbf{Object-Level}} & \multicolumn{2}{c}{\textbf{Attribute-Level}} & \textbf{Total Scores} \\
\cmidrule(lr){3-4} \cmidrule(lr){5-6}
 &  & {\textit{Existence}} & {\textit{Count}} & {\textit{Position}} & {\textit{Color}} &  \\
\midrule
\multirow{3}{*}{miniGPT4} & \textit{default} & 46.67 & 26.67 & 38.33 & 38.33 & 150.00 \\
 & \textit{+vcd} & 48.33 & 31.67 & 40.00 & 45.00 & 165.00 \\
 & \textit{+icd} & \textbf{66.67} & \textbf{61.67} & \textbf{40.00} & \textbf{61.67} & \textbf{230.01} \\
\addlinespace 
\multirow{3}{*}{InstructBLIP} & \textit{default} & 135.00 & 53.33 & 56.67 & 93.33 & 338.33 \\
 & \textit{+vcd} & 123.33 & 81.67 & 55.00 & 106.67 & 366.67 \\
 & \textit{+icd} & \textbf{136.67} & \textbf{90.00} & \textbf{76.67} & \textbf{123.33} & \textbf{426.67} \\
\bottomrule
\end{tabular}
}
\caption{\textbf{Results on the MME hallucination Subset.} The best performances within each setting are \textbf{bolded}.}
\label{tab:MME-Sub}
\end{table}

As detailed in Table~\ref{tab:MME-Sub}, our ICD method significantly surpasses the baseline LVLMs and the VCD method across all four tasks, demonstrating its superior capability in suppressing both object and attribute-level hallucinations with a large margin ($\textbf{+84.2}$ and $\textbf{+62.5}$ respectively in total scores). Interestingly, while the VCD method experiences a decline in performance on the \textit{position} hallucination task, our method maintains robust performance. This distinction underscores the adaptability and effectiveness of the ICD method in addressing a broader spectrum of hallucination symptoms, making it a more versatile solution in LVLMs.



\textbf{Results on MME Benchmark:} Our method is designed to mitigate hallucinations in LVLMs during inference. We delve deeper into ascertaining whether our approach not only preserves but potentially enhances the fundamental \textit{recognition} and \textit{reasoning} capabilities of LVLMs. To this end, we analyze performance across the full comprehensive MME benchmark, which encompasses 14 subtasks designed to assess \textit{perception} and \textit{recognition}.

Figure~\ref{fig:mme} illustrates that implementing ICD with both backbone models significantly improves task scores, surpassing the performance of foundation LVLMs and established VCD method. This outcome suggests that our method not only manages hallucinations effectively during inference but also elevates the accuracy of foundational LVLM tasks. 

In a more detailed model-specific analysis, our approach consistently outperforms both the backbone miniGPT4 and the VCD method with the same backbone across all 14 subtasks. Conversely, the VCD method exhibits diminished performance in specific areas such as \textit{posters, artwork, OCR, numerical calculation, text translation, and code reasoning} when compared to the baseline LVLM. 

Moreover, when InstructBLIP serves as the backbone, the effectiveness of VCD decreases in tasks related to \textit{existence, position, scene, and code reasoning}. We surmise that while leveraging visual uncertainty may anchor predictions more firmly in visual input, it simultaneously introduces drawbacks by fostering an over-reliance on visual cues at the expense of instruction-based grounding. Conversely, our ICD method, by focusing on multimodal alignment, does not compromise the fundamental reasoning capabilities of LVLMs. Notably, our method's performance on the \textit{landmark, OCR, commonsense reasoning, and text translation} tasks under InstructBLIP is weaker than the VCD method, whereas VCD exhibits superior results in these domains. This suggests that these subtasks within the MME benchmark may demand a robust visual discrimination capability.


\subsection{Discussions on ICD and VCD}
\label{sec:disicdvcd}
In addressing hallucinations in LVLMs, our ICD method and the baseline VCD both leverage contrastive decoding tailored for open-ended generation \cite{li2022contrastive}. While our ICD method introduces disturbance instructions to increase multimodal alignment uncertainty, VCD employs distorted images to amplify visual uncertainty. Positing that a synergistic approach could harness the strengths of both methods, we propose to analyze a straightforward integration of these two methods.

\begin{figure}[htbp]
  \centering
  \includegraphics[width=1.0\columnwidth]{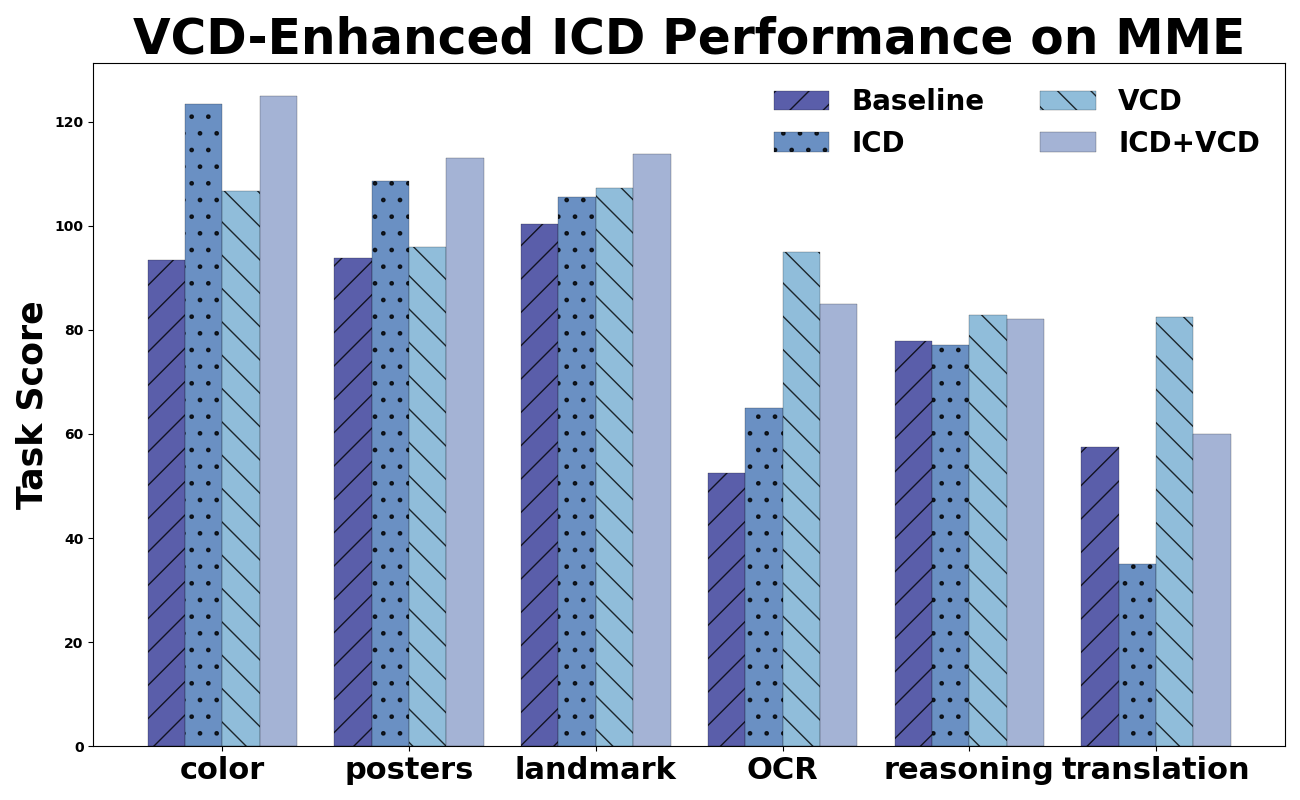}
  \caption{\textbf{Performance of the VCD-enhanced ICD method on MME Subset.} The underlying LVLM is InstructBLIP.}
  \label{fig:vcdicd} 
\end{figure}

Our combined approach begins with the VCD, utilizing standard instructions. This is followed by contrasting the resulting distribution with that of a VCD output generated under disturbance instructions, thereby establishing the final output distribution. Figure~\ref{fig:vcdicd} showcases the integration method on \textit{color, posters, landmarks, OCR, commonsense reasoning, and text translation}. This approach yields notable enhancements across these subtasks, underlining the importance of discriminative visual features and multimodal alignment as complements in grounding LVLM responses.

This exploration suggests a promising avenue for future research aimed at optimally amalgamating the advantages of both methods. Detailed results and comprehensive analysis of the combined method performance across full MME are provided in the appendix~\ref{sec:further} for further reference.


\section{Conclusion}
We introduce a novel instruction contrastive decoding approach that effectively detaches hallucinatory concepts by contrasting distributions derived from standard and disturbance instructions where role prefixes are appended to amplify hallucinations. Comprehensive experiments across various benchmarks and different LVLMs demonstrate the capability of our method in mitigating hallucinations and substantially improving the general perception and recognition performance of LVLMs.


\section*{Limitations}
In this paper, we have concentrated on addressing hallucinations within LVLMs by deploying our novel ICD method. We have validated its efficacy through rigorous evaluation on various hallucination discrimination benchmarks and have also qualitatively assessed its performance on generative benchmarks, which are pivotal for examining hallucinatory content. Despite their importance, generative benchmarks currently lack established metrics for thoroughly analyzing hallucinations, indicating a significant area for future research to enhance open-ended generation performance evaluation with robust automatic metrics.

\section*{Ethics Statement}
We propose the Instruction Contrastive Decoding method to address hallucination issues in LVLMs, thereby enhancing their safety and reliability within the community. Additionally, the datasets utilized for inferring and evaluating the ICD method are publicly accessible, promoting transparency and reproducibility in our research. Furthermore, we have made our code available to the public, ensuring it is convenient for researchers and practitioners to access and implement.

\section*{Acknowledgments}
We thank the anonymous reviewers and the area chair for their insightful comments and suggestions. This research was funded by the German Research Foundation DFG Transregio SFB 169: Crossmodal Learning: Adaptivity, Prediction, and Interaction.
\bibliography{anthology,custom}
\bibliographystyle{acl_natbib}

\appendix

\section{Implementation Details}
\label{sec:imdetail}

In our experiments, we adopted the contrastive decoding configurations by setting the decisive penalty on the decision made by LVLMs with disturbance $\lambda=1$ and the hyperparameter $\alpha=0.1$ that modulates the truncation of the probability distribution, in line with the configurations reported in previous studies \cite{li2022contrastive, leng2023mitigating}. For the decoding strategy, we uniformly applied the sampling method across all experiments, incorporating a $top$ $p = 1$, a $repetition$ $penalty = 1$, and a $number$ $of$ $beams = 1$ for LLMs. For both VCD and ICD methods, we sample from the modified $softmax$ distribution, as delineated in Equation~\ref{eq:myequation}.

We conducted experiments with various instructional disturbances, incorporating both positive and negative role prefixes. For illustrative purposes, we present two positive role prefix instructions and two negative role prefix instructions, providing a detailed guide for others to effectively implement our method in practical applications.

\begin{enumerate}
  \item[$\square$] \colorbox{pink}{\parbox{\dimexpr\linewidth-2\fboxsep\relax}{P1: \textit{You are an object detector to recognize every different object.}}}
  \item[$\square$] \colorbox{pink}{\parbox{\dimexpr\linewidth-2\fboxsep\relax}{P2: \textit{You are an object detector to recognize every different object by focusing on the shapes, colors, and relationships of objects.}}}
  \item[$\triangle$] \colorbox{lime}{\parbox{\dimexpr\linewidth-2\fboxsep\relax}{N1: \textit{I want you to avoid any specific identification or categorization of the objects depicted.}}}
  \item[$\triangle$] \colorbox{lime}{\parbox{\dimexpr\linewidth-2\fboxsep\relax}{N2: \textit{You are a confused object detector to provide a fuzzy overview or impression of the image.}}}
\end{enumerate}


\section{Qualitative Evaluation on LLaVa-Bench}
\label{sec:case}
\begin{figure*}[htbp] 
  \centering
  \includegraphics[width=0.8\textwidth]{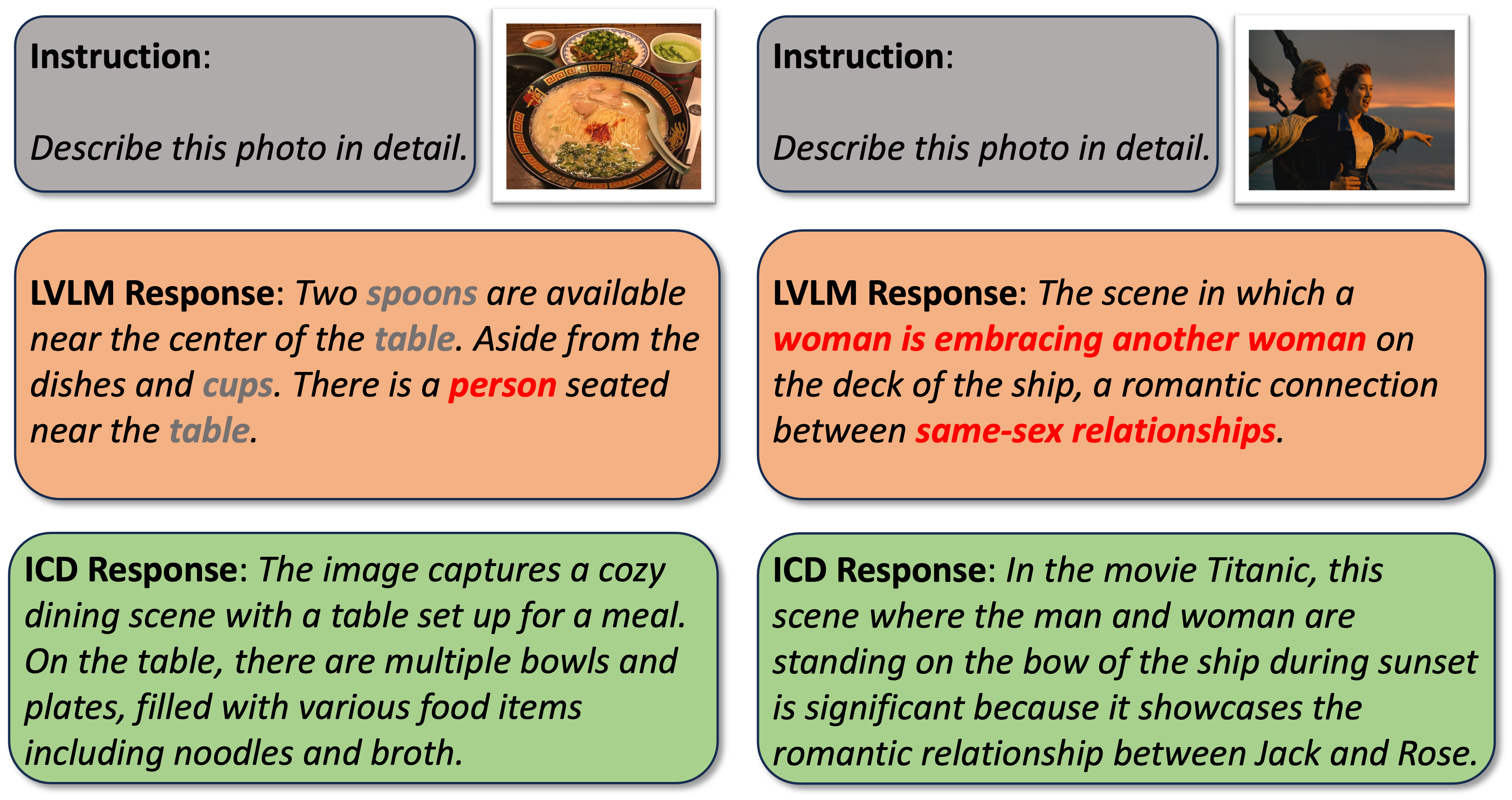} 
  \caption{\textbf{Qualitative analysis on LLava-Bench.} The left figure highlights the statistical bias, and the right figure shows the language prior that contributes to hallucinations in LVLMs. Hallucinated concepts have been highlighted in \textcolor{red}{\textbf{red}}.}
  \label{fig:llavacase} 
\end{figure*}

In this section, we extend our analysis by focusing on the evaluation of generative hallucination. Utilizing LLaVa-Bench, we conduct a qualitative analysis on the task of open-ended generation. Figure \ref{fig:llavacase} showcases two case studies that compare our method with backbone LVLMs using identical input images. The example displayed on the left presents various Asian dishes. While the baseline LVLMs accurately identify and generate concepts such as \textit{spoons, tables, and cups}, they also erroneously introduce the unrelated concept of a “\textit{person}.” This error stems from the high frequency of co-occurrence between \textit{“person” and "tables"} in the training data. Furthermore, the example on the right depicts a well-known scene from the movie "Titanic." Here, the baseline LVLMs incorrectly perceive the characters Jack and Rose as \textit{two women}, leading to an inaccurate generation of text regarding \textit{same-sex relationships}. This error is a result of the language prior biases, which contribute to hallucinations in LVLMs.

Contrastingly, our ICD approach produces fluent, coherent text that is closely grounded in the visual context, effectively mitigating the hallucinations caused by statistical biases and the inherent language priors of LVLMs.


\section{General QA Benchmarks Performance}
\label{sec:GQA}

Our evaluation in Section~\ref{experiment}, utilizing the POPE, MME, and LLava-Bench benchmarks, focused on assessing hallucinations in LVLMs, particularly regarding the presence of objects and attributes within images. In this section, we broaden our evaluation scope to include more general QA and caption datasets, such as MSCOCO \cite{lin2014microsoft}, OK-VQA \cite{marino2019ok}, and TextVQA \cite{singh2019towards}, using metrics like CHAIR \cite{rohrbach2018object}, CIDEr \cite{vedantam2015cider}, and BLEU \cite{papineni2002bleu} for a comprehensive analysis.

Specifically, MSCOCO comprises 118,000 images in the training set and 5,000 images in the validation set. Following the evaluation settings of \citet{rohrbach2018object}, we report instance-level (\textit{CHAIR\_I}) and sentence-level (\textit{CHAIR\_S}) hallucination results using the MSCOCO validation set in Table ~\ref{tab:mscoco-chairval}. Additionally, in line with \citet{yue2024less}, we randomly sample 500 instances from the MSCOCO training and validation sets, with our results presented in Table ~\ref{tab:mscoco-chairtrain}.

\begin{table}[h!]
\centering
\resizebox{0.4\textwidth}{!}{%
\begin{tabular}{lcc}
\toprule
\textbf{Method} & \textbf{CHAIR\_I $\downarrow$} & \textbf{CHAIR\_S $\downarrow$} \\
\midrule
InstructBLIP & 10.7 & 20.0 \\
VCD & 9.3 & 18.2 \\
\textbf{ICD} & \textbf{8.0} & \textbf{15.2} \\
\bottomrule
\end{tabular}
}
\caption{\textbf{Evaluation on MSCOCO validation set} using metrics CHAIR\_I and CHAIR\_S, instance-level and sentence-level hallucinations, followed by \citet{rohrbach2018object}.}
\label{tab:mscoco-chairval}
\end{table}

\begin{table}[h!]
\centering
\resizebox{0.4\textwidth}{!}{%
\begin{tabular}{lcc}
\toprule
\textbf{Method} & \textbf{CHAIR\_I $\downarrow$} & \textbf{CHAIR\_S $\downarrow$} \\
\midrule
InstructBLIP & 12.2 & 21.4 \\
VCD & 9.0 & 16.6 \\
\textbf{ICD} & \textbf{7.7} & \textbf{14.4} \\
\bottomrule
\end{tabular}
}
\caption{\textbf{Evaluation on MSCOCO training and validation sets (500 samples)}, using metrics CHAIR\_I and CHAIR\_S, instance-level and sentence-level hallucinations, followed by \citet{yue2024less}.}
\label{tab:mscoco-chairtrain}
\end{table}

We further evaluate our ICD method on the OK-VQA and TextVQA datasets using CIDEr and BLEU metrics, with the improved results detailed in Tables ~\ref{tab:ok-vqa} and ~\ref{tab:textvqa}.

\begin{table}[h!]
\centering
\resizebox{0.42\textwidth}{!}{%
\begin{tabular}{lccc}
\toprule
\textbf{Method} & \textbf{CIDEr} & \textbf{BLEU1} & \textbf{BLEU2} \\
\midrule
InstructBLIP & 0.28 & 0.33 & 0.17 \\
VCD & 0.35 & 0.42 & 0.22 \\
\textbf{ICD} & \textbf{0.40} & \textbf{0.45} & \textbf{0.25} \\
\bottomrule
\end{tabular}
}
\caption{\textbf{Evaluation on OK-VQA test set} using metrics CIDEr and BLEU 1 and 2.}
\label{tab:ok-vqa}
\end{table}

\begin{table}[h!]
\centering
\resizebox{0.48\textwidth}{!}{%
\begin{tabular}{lccccc}
\toprule
\textbf{Method} & \textbf{CIDEr} & \textbf{BLEU1} & \textbf{BLEU2} & \textbf{BLEU3} & \textbf{BLEU4} \\
\midrule
InstructBLIP & 0.56 & 0.29 & 0.22 & 0.19 & 0.19 \\
\textbf{VCD} & \textbf{0.71} & \textbf{0.36} & \textbf{0.32} & \textbf{0.30} & \textbf{0.31} \\
\textbf{ICD} & \textbf{0.69} & \textbf{0.35} & \textbf{0.30} & \textbf{0.29} & \textbf{0.30} \\
\bottomrule
\end{tabular}
}
\caption{\textbf{Evaluation on TextVQA test set} using metrics CIDEer and BLEU 1, 2, 3, and 4.}
\label{tab:textvqa}
\end{table}

These additional evaluations underscore the versatility and efficacy of the ICD method across diverse QA datasets and evaluation metrics. Notably, the improvements in metrics such as CHAIR, CIDEr, and BLEU across the MSCOCO, OK-VQA, and TextVQA benchmarks reaffirm the significant impact of our method.

\section{Further Analysis on VCD-Enhanced ICD}
\label{sec:further}

\begin{figure*}[htbp] 
  \centering
  \includegraphics[width=0.7\textwidth]{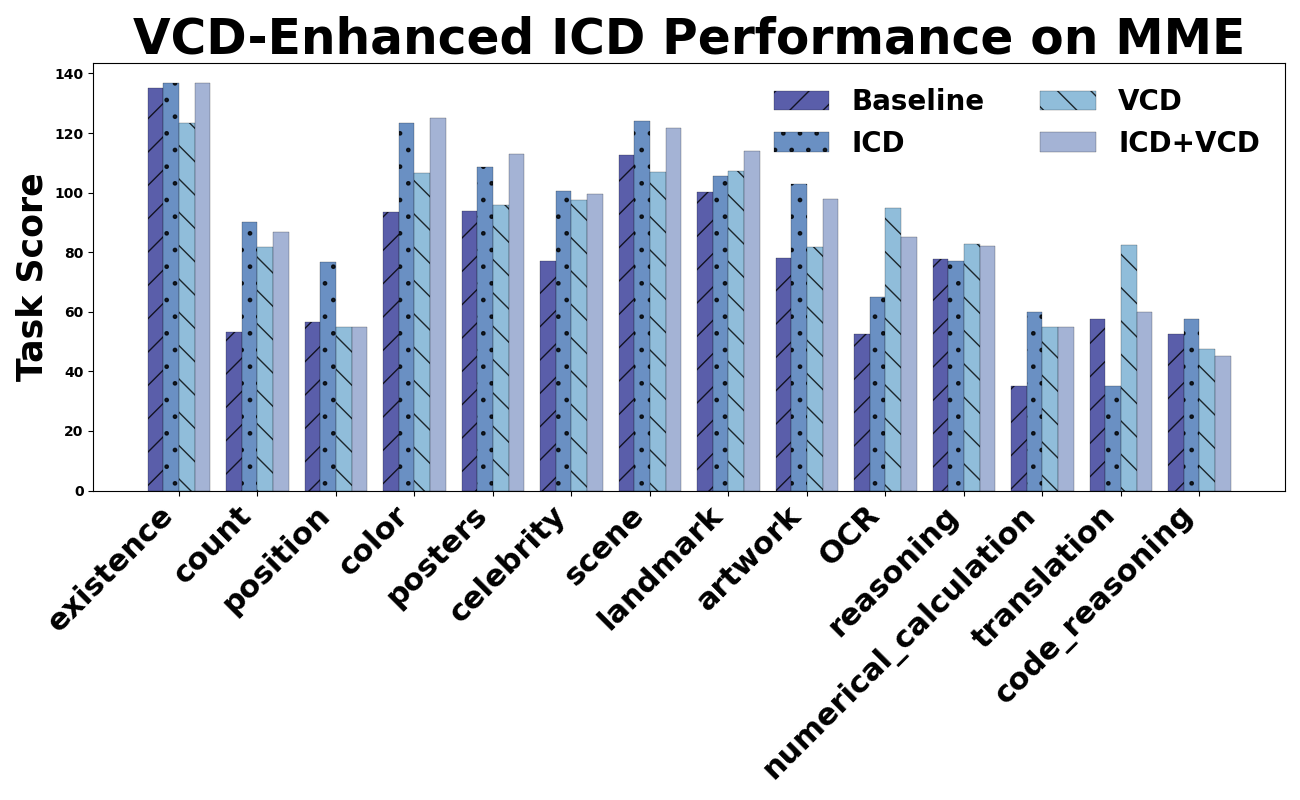} 
  \caption{\textbf{Performance of the VCD-enhanced ICD method on full MME benchmark.} The underlying LVLM is InstructBLIP. ICD+VCD indicates the combination approach detailed in Section~\ref{sec:disicdvcd}.}
  \label{fig:addpic} 
\end{figure*}

We comprehensively analyze the ICD and VCD combined method, detailed in Section~\ref{sec:disicdvcd}, within the full MME benchmark, utilizing InstructBLIP as the backbone LVLM. Figure~\ref{fig:addpic} illustrates that integrating our ICD method significantly enhances the VCD's performance across various tasks, including \textit{existence, count, color, celebrity, scene, landmark, and artwork}. Similarly, incorporating VCD in ICD yields improvements in \textit{color, posters, landmarks, OCR, commonsense reasoning, and translation tasks}. These findings suggest that addressing both visual and multimodal alignment uncertainties in a complementary fashion effectively mitigates hallucinations. However, we also note a performance decrement in the ICD method for \textit{count, position, artwork, calculation, and code reasoning tasks} when combined with VCD. This observation underscores the necessity for more refined combination strategies to fully harness the potential of integrating these two methods.

Combining the strengths of both the ICD and VCD methods has opened a promising avenue for future investigations. We aim to develop and refine contrastive decoding methods for the seamless integration of both techniques, potentially a new method for mitigating hallucinations in LVLMs.

\section{Optimal Position to Apply Contrastive Decoding}
\label{sec:cd}

\begin{figure}[htbp]
  \centering
  \includegraphics[width=1\columnwidth]{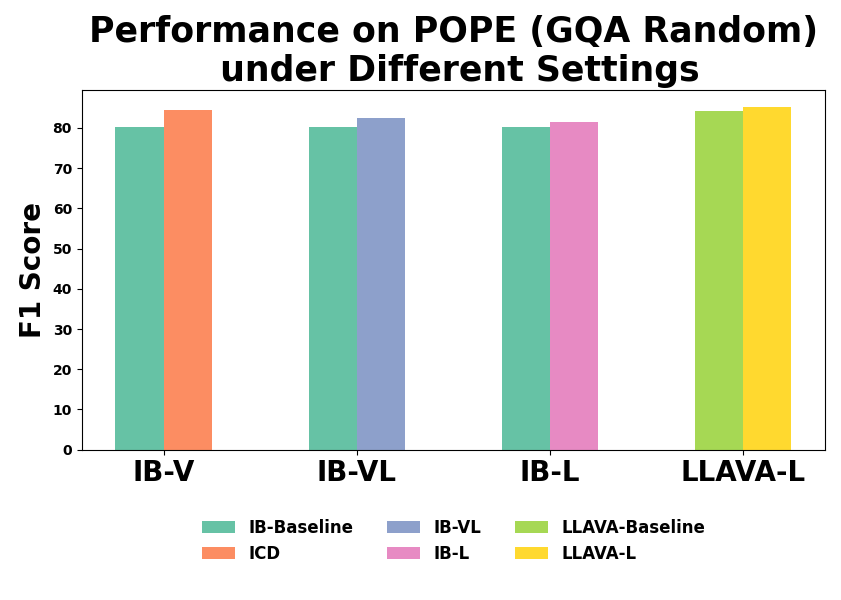}
  \caption{\textbf{Performance of the ICD method implemented on difference positions evaluated on POPE (GQA Random) dataset.} The underlying LVLMs are InstructBLIP and LLaVa-1.5.}
  \label{fig:postionof} 
\end{figure}

Upon detailed examination of the inference framework depicted in Figure~\ref{fig:pnprefix}, we identify three potential points for integrating the ICD method: within the Q-Former's instruction, the LLM's instruction, and a combination of both. This analysis, based on the POPE (\textit{GQA random setting}), aims to pinpoint the optimal implementation site for ICD. To ensure a comprehensive comparison, we selected two distinct LVLMs, InstructBLIP and LLaVa, as backbones to represent varied fusion approaches. InstructBLIP employs Q-Former for multimodal alignment, whereas LLaVa utilizes a linear projection.

Figure~\ref{fig:postionof} reveals that, under the InstructBLIP framework, ICD enhances performance across all implementation sites, with the singular application within Q-Former yielding the most significant improvement. A comparison between the LVLMs indicates that LLaVa also benefits from the ICD method when ICD is applied within LLMs. However, exclusive application of ICD in LLMs produces less pronounced improvements, mirroring the observations with InstructBLIP as the backbone. Consequently, our findings suggest that deploying the ICD method within the Q-Former architecture represents the most effective strategy.

\end{document}